\documentclass{article}

\PassOptionsToPackage{numbers, compress}{natbib}
\usepackage[final]{neurips_2020}

\usepackage[utf8]{inputenc} 
\usepackage[T1]{fontenc}    
\usepackage{hyperref}       
\usepackage{url}            
\usepackage{booktabs}       
\usepackage{amsfonts}       
\usepackage{nicefrac}       
\usepackage{microtype}      
\usepackage{amsmath,booktabs}

\usepackage{graphicx}
\usepackage{savesym}
\usepackage[linesnumbered, ruled,vlined]{algorithm2e}
\savesymbol{listofalgorithms}
\usepackage{algorithm,algorithmic}
\usepackage{tikz}
\usetikzlibrary{positioning}
\usetikzlibrary{intersections,shapes.arrows}
\usepackage{filecontents}
\usepackage{epsdice}
\usepackage{hyperref}
\usepackage{bm}
\usepackage{wrapfig}
\usepackage{multirow}

\usepackage{todonotes}

\usepackage{url}
\usepackage{graphicx}
\usepackage[font=scriptsize]{caption}
\usepackage{amsmath}

\usepackage{amsfonts}
\usepackage{graphicx} 
\usepackage{natbib}

\usepackage{xcolor}
\usepackage{tabularx} 
\usepackage{amsthm}
\usepackage{amsfonts}
\usepackage{amsmath}
\usepackage{amssymb}
\usepackage{mathrsfs}
\usepackage{multirow}
\usepackage{bbm}
\usepackage{wrapfig}

\usepackage{enumitem}
\usepackage{wrapfig}

\newcommand{\beq}{\begin{equation}}

\newcommand{\eeq}{\end{equation}}
\newcommand{\beqs}{\begin{equation*}}
\newcommand{\eeqs}{\end{equation*}}

\usepackage[framemethod=TikZ]{mdframed}
\usepackage{amsthm}
\newcounter{theo}[section] 
\renewcommand{\thetheo}{\arabic{section}.\arabic{theo}}

\newcounter{lem}[section] 
\renewcommand{\thelem}{\arabic{section}.\arabic{lem}}

\newcounter{prf}[section]
\renewcommand{\theprf}{\arabic{section}.\arabic{prf}}

\newcounter{cor}[section]
\renewcommand{\thecor}{\arabic{section}.\arabic{cor}}

\newcounter{prop}[section] 
\renewcommand{\thelem}{\arabic{section}.\arabic{prop}}

\renewcommand{\algorithmiccomment}[1]{\bgroup\hfill//~#1\egroup}

\usepackage{comment}

\usepackage{caption}
\title{Learned Equivariant Rendering without Transformation Supervision}

\author{%
  Cinjon Resnick\thanks{Correspondence to cinjon@nyu.edu.} \\
    NYU \\
  \And
  Or Litany \\
  NVidia \\
  \And
  Hugo Larochelle \\
  Google \\
  \And
  Joan Bruna \\
  NYU \\
  \And
  Kyunghyun Cho \\
  NYU 
}

\begin{document}

\maketitle

\begin{abstract}

We propose a self-supervised framework to learn scene representations from video that are automatically delineated into objects and background. Our method relies on moving objects being equivariant with respect to their transformation across frames and the background being constant. 
After training, we can manipulate and render the scenes in real time to create unseen combinations of objects, transformations, and backgrounds. We show results on moving MNIST with backgrounds.


\end{abstract}

\section{Introduction}
\label{sec:intro}

Learning manipulable representations of scenes is a challenging task. Ideally, we would give our model an image of a scene and receive an inverse rendering of coherent objects along with a static background. This is infeasible without an inductive bias because the problem is ill-posed. For example, distinguishing a previously unseen object from the background is inherently ambiguous. Consequently, prior works introduce mechanisms to help the model at training or inference time. We do similarly and choose a mechanism that intuitively follows from learning from video.

Assume an input video of a dynamic scene captured by a static camera. The static background is the implicit objects that are constant across the frames\footnote{Note that with this definition of the background, occlusions are just occasions when the implicit object is not visibly painted in the scene. Practically, this can be handled with alpha transparency.}. In contrast, the foreground is the implicit objects that are equivariant with respect to a family of transformations containing more than just the identity. We use this difference to infer a scene representation that captures separately the background and the moving object.  
Like \citet{dupont2020equivariant}, we limit the family of transformations to be affine. Building on their work, we attain two novel results:

\begin{enumerate}
    \item \textbf{Learn the transformation}: We learn the tranformation from nearby video frames and do not require it as input during training. This has further benefits at inference time. 
    \item \textbf{Distinguish objects and background}: By encoding the background as the constant objects and the manipulable character as the equivariant object, we yield orthogonal encoders for the character and the background and can consequently manipulate them independently.
\end{enumerate}

Our model is trained in a self-supervised fashion with only rendered pairs $(x^{i}_p, x^{i}_q)$ of nearby frames from video sequence $i$. We impose no other constraint and can render new scenes combining objects and backgrounds in real time with additional (potentially unrelated) frames $x^{j}$ and $x^{k}$. We show strong results in Sec~\ref{sec:experiments} on a 2D task involving moving MNIST\cite{lecun-mnisthandwrittendigit-2010} digits on static backgrounds where we demonstrate the following manipulations:

\begin{itemize}
    \item Render the object in $x^{i}_1$ but with the background from $x^{j}_1$.
    \item Render the object in $x^{i}_1$ using the transformation seen from $x^{j}_p \rightarrow x^{j}_q$.
    \item Combine the above two manipulations to render the object in $x^{i}_1$ on the background from $x^{j}_1$ using the transformation exhibited by the change seen from $x^{k}_p \rightarrow x^{k}_q$.
\end{itemize}


\section{Background}
\label{sec:background}

\citet{dupont2020equivariant} denoted an image $x \in X = \mathbb{R}^{c \times h \times w}$ along with a scene representation $z \in Z$, a rendering function $g: Z \rightarrow X$ mapping scenes to images, and an inverse renderer $f: X \rightarrow Z$ mapping images to scene representations. It is helpful to consider $x$ as a 2D rendering of an implicit object $o$ from a specific camera viewpoint. Then, with respect to transformation $T$, an equivariant scene representation $z$ satisfies the following relation:
\begin{equation}
\label{eqn:equivariant}
    T^{X}g(z) = g(T^{Z}z)
\end{equation}

\begin{wrapfigure}{r}{0.35\textwidth} 
    \centering
    \includegraphics[width=.35\textwidth]{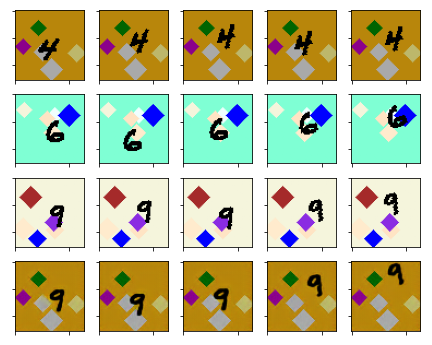}
    \caption{\textbf{Inference demonstration}: The first three rows are ground truth from the test set. The fourth row is the object from the third row, the transformations in the second row, and the background from the first row.}
    \label{fig:manipulable}
\end{wrapfigure}

In other words, transforming a rendering in image space is equivalent to first transforming the scene in feature space and then rendering the result. Here, $x$ is a 2D image rendering, $z$ is that rendering's scene representation, and $T^Z$ is an affine transformation. We then learn neural networks $f$ and $g$ s.t.:
\begin{equation}
\label{eqn:autoencoder}
    x_2 = T^{X}x_1 = T^{X}g(z_1) = g(T^{Z}z_1) = g(T^{Z}f(x_1))
\end{equation}

In \cite{dupont2020equivariant}, they require triples $(x_1, x_2, \theta)$. The pair $x_1$ and $x_2$ are renderings of the same object $o$, and $\theta$ is the rotation transforming $o$ from its appearance in $x_1$ to its appearance in $x_2$. They use the same $\theta$ for $T^Z$, which means that the 3D representation $z_1 = f(x_1)$ is rotated by $\theta$. This yields $\tilde{z}_1 = R^{Z}_{\theta}f(x_1)$ and $\tilde{z}_2 = (R^{Z}_{\theta})^{-1}f(x_2)$, with which they train $g$ and $f$ to minimize reconstruction loss (Eqn~\ref{eqn:loss}):
\begin{align}
    \mathbb{L}_{\text{render}} = ||x_2 - g(\tilde{z_1})|| + ||x_1 - g(\tilde{z_2})||
    \label{eqn:loss}
\end{align}

After training, they infer new renderings by first inverse rendering $z = f(x)$, then applying rotations $T^Z$ upon $z$, and finally rendering images $\hat{x} = g(T^{Z}z)$. This is notable because it means we can operate entirely in feature space. As we show in Sec~\ref{sec:method}, this lets us manipulate the output rendering in ways that are very difficult to perform in image space.

\section{Method}
\label{sec:method}

Our motivation is in asking if we can we use the smoothly-changing nature of video to learn the transformations between frames. Like \cite{dupont2020equivariant}, we assume that the change between frame $F_t$ and $F_{t+1}$ can be modeled with affine transformations. However, while they use one (rotation) transformation, we assume an arbitrary affine transformation on the character object plus an invariant transformation on a static background (in-painting as needed). Additionally, we remove the $\theta$ requirement at both training and test by learning the transformation $T^Z$ from data. 

\paragraph{Learning the transformation} That video changes smoothly lets us advance past affine transformations parametrically defined with a $\theta$ angle of rotation and instead learn transformations based on frame changes. One advantage is that the model is now agnostic to which affine transformations the data exhibits. Another is that at inference time, the model is agnostic to whether the frame to be transformed is input to the transformation function.

\paragraph{Distinguishing objects and background} The renderer $g$ is equivariant to moving objects. A static background however will be constant across a video. We take advantage of this to learn $g$ along with functions $f_o$ and $f_b$ corresponding to the moving object and the background such that, at inference time, we can mix and match objects and backgrounds that we have previously never seen together.

\paragraph{Setup} Building on Sec~\ref{sec:background}, we define $f_b$ and $f_o$ respectively for the encoding of the background and the object. Following Eqn~\ref{eqn:autoencoder}, we then learn neural networks $f_o$, $f_b$, $g$, and $T^Z$ such that:
\begin{align}
    x_2 = g\left(T^{Z}(f_{o}(x_1), f_{o}(x_2)) \circ f_{o}(x_1) + f_{b}(x_1)\right)
\end{align}

During training, we require only the pair $(x_1, x_2)$. We find that two more constraints help. The first is that the object encoder is equivariant with respect to the transformation. The second is that the background encoder is constant. These are described below where we optimize $\mathbb{L}_{\text{total}}$ with scalar coefficients $\alpha_{\text{equiv}}, \alpha_{\text{inv}}$:
\begin{align}
    \mathbb{L}_{\text{scene}} &= ||g\left(T^{Z}(f_{o}(x_1), f_{o}(x_2)) \circ f_{o}(x_1) + f_{b}(x_1)\right) - x_2||_2 \\
    \mathbb{L}_{\text{equiv}} &= ||T^{Z}(f_{o}(x_1), f_{o}(x_2)) \circ f_{o}(x_1) - f_{o}(x_2)||_2 \\
    \mathbb{L}_{\text{inv}} &= ||f_b(x_1) - f_b(x_2)||_2 \\
    \mathbb{L}_{\text{total}} &= \mathbb{L}_{\text{scene}} + \alpha_{\text{equiv}}\mathbb{L}_{\text{equiv}} +
    \alpha_{\text{inv}}\mathbb{L}_{\text{inv}}
\end{align}



With this setup, both $T^Z$ and $f_o$ learn to handle every object similarly. This is important because it means that at inference time we can render novel scenes given a pair of nearby frames $(x_1, x_2)$ in a video. Denoting $h(x_1, x_2, x_3, x_4) = g\left(T^{Z}(f_{o}(x_1), f_{o}(x_2)) \circ f_{o}(x_3) + f_{b}(x_4)\right)$, the novel renderings described in Sec~\ref{sec:intro} and shown in Sec~\ref{sec:experiments} are:

\begin{itemize}
    \item $\bf{h(x^{i}_1, x^{i}_2, x^{i}_1, x^{j}_1)}$: Render the object in $x^{i}_1$ as it is in $x^{i}_2$ but with the background from $x^{j}_1$.
    \item $\bf{h(x^{i}_1, x^{i}_2, x^{j}_1, x^{j}_1)}$: Render the object in $x^{j}_1$ using the transformation exhibited by the change in the object from $x^{i}_1 \rightarrow x^{i}_2$.
    \item $\bf{h(x^{i}_1, x^{i}_2, x^{j}_1, x^{k}_1)}$:
    Combine the above two to render the object in $x^{j}_1$ on the background from $x^{k}_1$ using the transformation exhibited by the change in the object from $x^{i}_1 \rightarrow x^{i}_2$.
\end{itemize}


\section{Experiments}
\label{sec:experiments}

We show experiments on a dataset built on MNIST. This test-bed is suitable for demonstrating that our method is capable of both learning the transformations and separating objects from the background.

\paragraph{Dataset} We generate videos, each of length $M=5$, of MNIST digits (objects) moving on a static background. The digits and background have dimensions $(28, 28)$ and $(64, 64)$ respectively. At each training step, we select $N$ digits in the train split of MNIST, as well as a background from the set of pre-generated training backgrounds (see below). We then place these digits at some random initial position. For each digit, and for each of $M-1$ times, we choose randomly between either rotation or translation. If we choose translation, then we translate the object independently in each of the $x$ and $y$ directions by some random amount in $[-10, -8, \ldots, 8, 10] \setminus {0}$. If we choose rotation, then we rotate the object by some random amount in $[-15, -12, \ldots, 12, 15] \setminus {0}$. In both cases, if character leaves the boundaries of the image, then we redo the transformation selection. Otherwise, that transformation is applied cumulatively to yield the next object position. 


At this point, we have MNIST images on blank canvases. We overlay them on the chosen background to produce a sequence of images where the change in each object from frames $F_T \rightarrow F_{T+1}$ is small and affine for the object and constant for the background. Afterwards, we randomly choose two indices $i, j$ and use $(x_i, x_j)$ as the training pair. See Figure~\ref{fig:manipulable} for example sequences.

\paragraph{Backgrounds} We create $64$ randomly generated backgrounds for each of train and test. For each background, we select a color from the Matplotlib CSS4 colors list. We then place five diamonds on the background, each with a different random color, along with an independent and randomly chosen center and radius. The radius is uniformly chosen from between seven and ten, inclusive. 


\begin{wrapfigure}{R}{0.35\textwidth} 
    \centering
    \includegraphics[width=.35\textwidth]{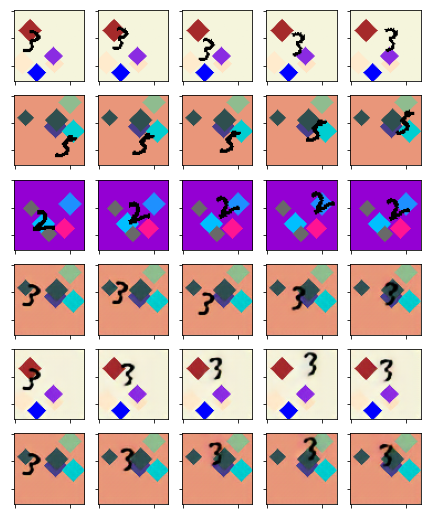}
    \caption{Section~\ref{sec:method} manipulations. With counterclockwise direction and bottom left origin, the transformations in the third row are rotate(15), rotate(9), translate(10, 4), translate(-8, -6).}
    \label{fig:allmanipulations}
\end{wrapfigure}

\paragraph{Model} We use neural networks for $f_o$, $f_b$, $g$, and $T^Z$. While distinct, both $f_o$ and $f_b$ share the same architecture details. The renderer $g$ is a transposition of that architecture, albeit without being residual. The transformation $T^Z$ has three important aspects. First, it is input-order dependent. Second, it uses PyTorch's\cite{paszke2017automatic} affine\_grid and grid\_sample functions to transform the scene similarly to how Spatial Transformers\cite{DBLP:journals/corr/JaderbergSZK15} operate. The third is that it is initialized at identity.

\paragraph{Qualitative Results} Our model learns to render new scenes using objects from the test set of MNIST as well as backgrounds it has never seen before. All shown sequences are on unseen backgrounds with unseen MNIST digits where there at least two transformations of each type (rotation and translation) and the transformations were cumulatively large over the sequence. We did not need to cherry-pick any of the results.

Figure~\ref{fig:allmanipulations} concisely demonstrates the manipulations from Sec~\ref{sec:method}. The first three rows are ground truth from the test set. The fourth row is replacing the background in the first row with that of the second row. The fifth row is applying the transformations in the third row to the first row's character and background. And the sixth row is both manipulations simultaneously. 

In particular, the final row is rendered by encoding the character from the first row, the background from the second row, and using the transformations exhibited in the third row. With $x^{i}_{j}$ as the $j$th frame from the $i$th sequence: $$x^{6}_{k} = g\left(T^{Z}(f_{o}(x^{3}_{1}), f_{o}(x^{3}_{k})) \circ f_{o}(x^{1}_{k}) + f_{b}(x^{2}_{k})\right)$$

\begin{wrapfigure}{R}{0.35\textwidth} 
    \centering
    \includegraphics[width=.35\textwidth]{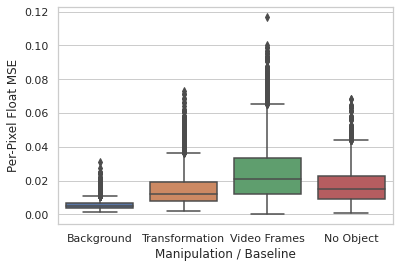}
\caption{Per-pixel MSE over 10,000 test examples. The transform and background manipulations use our learned functions; Video frames is MSE of a frame against a random (non-identical) frame from the same video; No object is MSE of the background versus the full frame of background and object.}
    \label{fig:pixelmse}
\end{wrapfigure}

\paragraph{Quantitative Results} We tested reconstruction results by evaluating the per pixel float MSE over the MNIST test set. For each example, we randomly chose two pairs of (background, digit) and made corresponding videos $(x^1_1, \ldots, x^1_5)$ and $(x^2_1, \ldots, x^2_5)$. We then indexed into the same random position in both sequences to get frame pairs $(x^1_i, x^1_j)$, $(x^2_i, x^2_j)$.

To get the MSE of transformation manipulations, we render the object and background from $x^2_{i}$ but transformed like in $x^1_{i} \rightarrow x^1_{j}$. We compare this ground truth to $\hat{x}^2_{j} = g(T^{Z}(f_{o}(x^1_{i}), f_{o}(x^1_{j})) \circ f_{o}(x^2_{i}) + f_{b}(x^2_{i})$. To get the MSE of background manipulations, we render the object in $x^1_{j}$ on the background from $x^2_{j}$ as ground truth and compare it to $\hat{x}^1_{j} = g(T^{Z}(f_{o}(x^1_{i}), f_{o}(x^1_{j})) \circ f_{o}(x^1_{i}) + f_{b}(x^2_{i})$. 

Fig~\ref{fig:pixelmse} shows a boxplot of these results along with two baselines: \textit{Video frames} is the MSE of two random frames from the same video; \textit{No object} is the MSE of a full frame against only the background from that frame. Given that MSE is a measure of reconstruction quality with lower values being better, we expect them to serve as upper bounds. \textit{Video frames} is the upper bound when reconstruction gets the object but places it incorrectly. \textit{No object} is the upper bound when reconstruction fails to include the object. On this measure, we see that the background manipulation is much better than the baselines, but we cannot say with certitude that the transform manipulation is better as it is within confidence interval of \textit{Video Frames} and its box plot overlaps with both baselines.


\section{Related Work}

Besides \cite{dupont2020equivariant}, two other related works are \citet{DBLP:journals/corr/abs-1710-07307,DBLP:journals/corr/abs-1904-06458}. They also rely on equivariance to learn representations capable of manipulating scenes. However, they do not delineate objects and backgrounds, nor do they learn the $T^Z$ from data. \citet{dupont2020equivariant} assumes that $T^Z$ is given during training; In \citet{DBLP:journals/corr/abs-1710-07307}, $T^Z$ is a block diagonal composition of (given) domain-specific transformations. \citet{DBLP:journals/corr/abs-1904-06458} uses a user-provided transformation. That we learn it from data lets us work with datasets where we do not have ground truth.

\citet{NIPS2015_5845} is also related. They apply transformations to frames to yield an analogous frame. However, they assume that the applying operation is addition rather than spatial transform, and they require an additional frame as input to $T^Z$ at inference (3) and two additional during training (4).


\section{Conclusion}

In this work, we have presented a framework for learning an equivariant renderer capable of delineating objects and the background such that it can manipulate each independently. Further, our framework only requires self-supervised video sequences and does not need labels.

Our assumption that the transformation between frames is affine does not hold in general; a contrary example is videos with nonlinear lighting effects. While there are real applications where it does occur such as stop-motion animation, relaxing this assumption is an area we are actively considering. A more pressing direction though is increasing the complexity of the datasets on which we experiment. We leave that extension to future work.


\bibliographystyle{plainnat}
\bibliography{refs}

\newpage
\appendix

\section*{Appendix}

\paragraph{Further examples} Here, we show more examples of our model manipulating sequences.

\begin{figure*}[ht]
    \centering
    \includegraphics[width=.6\textwidth]{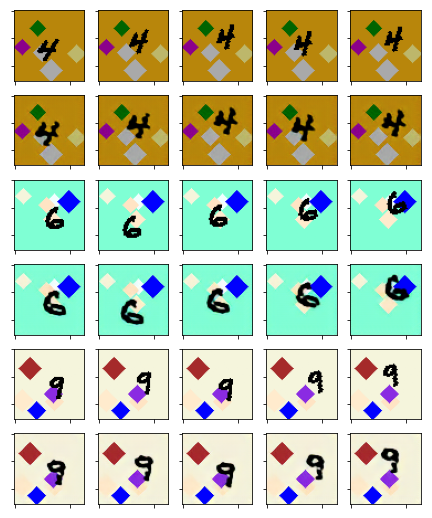}
    \caption{\textbf{Reconstructions}: The first, third, and fifth rows are original sequences. The second, fourth, and sixth rows are reconstructions of the prior row where $T^Z$ is fixed as the same transformation as $T^X$. Note that in these scenarios, the results are not as strong as when $T^Z$ is a learned function.}
    \label{fig:reconstruct}
\end{figure*}

\begin{figure*}[ht]
    \centering
    \includegraphics[width=.6\textwidth]{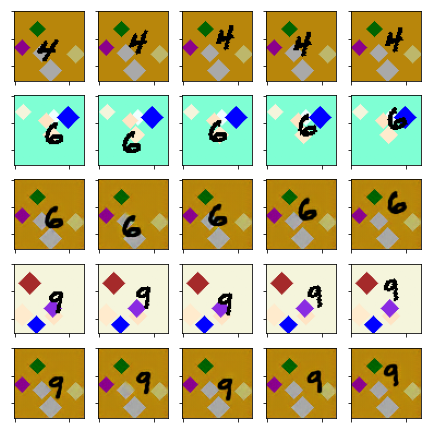}
    \caption{\textbf{Backgrounds}: The first, second, and fourth rows are originals. The third and fifth rows are the prior row but with the background changed to that of the first sequence.}
    \label{fig:background}
\end{figure*}

\begin{figure*}[ht]
    \centering
    \includegraphics[width=.6\textwidth]{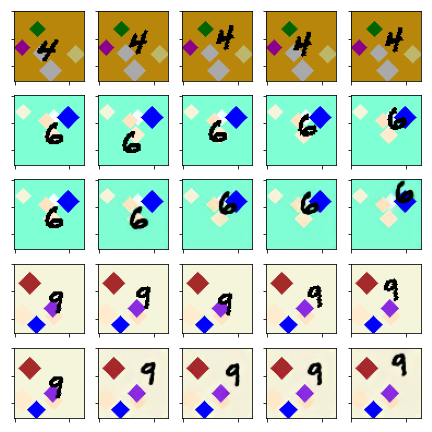}
    \caption{\textbf{Transformations}: The first, second, and fourth rows are originals. The third and fifth rows are the prior row but the transformation is a function of the first row.}
    \label{fig:transformation}
\end{figure*}

\clearpage

\paragraph{Analyzing $T^Z$} We compare the learned $T^Z$ to the ground truth object transformation. Each column in \ref{table:transform-matrices} shows a $2 \times 3$ matrix representing the independent statistics of each entry of the transformation in question. These statistics are attained over 40,000 unique frame pairs. For example, the first column shows the mean of each entry in the transformation matrix.

\begin{table}[htp]
\centering\footnotesize

\setlength{\tabcolsep}{0pt}
\addtolength{\arraycolsep}{-3pt}

\begin{tabular*}{\textwidth}{@{\extracolsep{\fill}}c c c c}
\toprule
Transformation & Mean & Max & Min \\
\midrule
Ground Truth &
$\begin{pmatrix}0.990 & -0.003 & 0.558 \\ 0.003 & 0.990 & 0.424 \\ \end{pmatrix}$ &
$\begin{pmatrix}1.000 & 0.500 & 26.74 \\ 0.588 & 1.000 & 30.00 \\ \end{pmatrix}$ &
$\begin{pmatrix}0.809 & -0.588 & -28.00 \\ -0.500 & 0.809 & -26.00 \\ \end{pmatrix}$ \\
\midrule
Learned $T^Z$ &
  $\begin{pmatrix}1.197 & 0.050 & -0.026 \\ 0.178 & 1.109 & -0.023 \\ \end{pmatrix}$ &
$\begin{pmatrix}1.061 & -0.324 & -1.066 \\ -0.063 & 1.026 & -1.157 \end{pmatrix}$ &
 $\begin{pmatrix}1.268 & 0.362 & 0.936 \\ 0.314 & 1.262 & 0.889 \\ \end{pmatrix}$ \\
\bottomrule
\end{tabular*} 
\caption{Independent statistics of the transformations.}
\label{table:transform-matrices}
\end{table}

\paragraph{Example background} The diamond colors and radii are randomly chosen and distinct.

\begin{figure*}[h]
    \centering
    \includegraphics{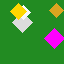}    
    \caption{Example background.}
    \label{fig:background_examples}
\end{figure*}

\paragraph{PSNR of Baselines and Manipulations} PSNR over 10,000 test examples. The transform and background manipulations are done with our learned functions. Video frames is the PSNR of a frame against a random (non-identical) frame from the same video. No object is the PSNR of the background versus the full frame of background plus object.

\begin{table}[!h]
\centering
\begin{tabular}{c|c}
Type & PSNR 95\% CI \\
\toprule
Background manipulation & $22.920 \pm .033$ \\
Transform manipulation & $18.912 \pm .051$ \\
Baseline: Video frames & $18.993 \pm 0.319$ \\
Baseline: No object & $18.609 \pm .060$ \\
\end{tabular}
\caption{PSNR.}
\label{table:psnr}
\end{table}

\end{document}